\documentclass{article}
\usepackage{./VLA-AN-style} 
\usepackage{graphicx}
\usepackage{amsmath}
\usepackage{amssymb}
\usepackage{booktabs}
\usepackage{hyperref}
\usepackage{multirow}
\usepackage{url}
\usepackage{makecell}
\usepackage{cite}

\title{} 
\author{} 
\date{} 

\begin{document}
\maketitle

\begin{abstract}
This paper proposes VLA-AN ($\mathbf{V}$ision-$\mathbf{L}$anguage-$\mathbf{A}$ction framework for $\mathbf{A}$erial $\mathbf{N}$avigation), an efficient and onboard Vision-Language-Action (VLA) framework dedicated to autonomous drone navigation in complex environments. VLA-AN addresses four major limitations of existing large aerial navigation models: the data domain gap, insufficient temporal navigation with reasoning, safety issues with generative action policies, and onboard deployment constraints. First, we construct a high-fidelity dataset utilizing 3D Gaussian Splatting (3D-GS) to effectively bridge the domain gap. Second, we introduce a progressive three-stage training framework that sequentially reinforces scene comprehension, core flight skills, and complex navigation capabilities. Third, we design a lightweight, real-time action module coupled with geometric safety correction. This module ensures fast, collision-free, and stable command generation, mitigating the safety risks inherent in stochastic generative policies. Finally, through deep optimization of the onboard deployment pipeline, VLA-AN achieves a robust real-time inference rate of 2–3 Hz on resource-constrained UAVs. Extensive experiments demonstrate that VLA-AN significantly improves spatial grounding, scene reasoning, and long-horizon navigation, achieving a maximum single-task success rate of 98.1\%, and providing an efficient, practical solution for realizing full-chain closed-loop autonomy in lightweight aerial robots.
\end{abstract}


\section{Introduction}

The rapid and concurrent advances of Multimodal Large Language Models (MLLMs) \cite{liu2023visualinstructiontuning,openai2023gpt4,team2023gemini,bai2023qwenvl} have fundamentally reshaped the imagination for the intelligence and autonomy of Unmanned Aerial Vehicles (UAVs). The emerging paradigm envisions aerial systems endowed with human-like cognitive–action capabilities, such as interpreting complex natural language instructions, understanding fine-grained scene semantics, inferring nuanced task intentions, and generating context-aware action sequences. However, most existing UAV systems remain largely dependent on manual control or limited autonomy. Their designs typically follow a cascaded perception–mapping–planning–control pipeline, composed of carefully engineered modules tailored to specific tasks such as tracking \cite{9811688}, landing \cite{9981489}, or physical interaction \cite{10044964}. While effective in constrained settings, this modular paradigm introduces two fundamental limitations. First, errors tend to accumulate across stages, degrading overall system robustness. Second, these systems lack the capacity to reason over open-ended language and high-level intent. Extending them to new tasks often requires substantial manual redesign, making adaptation slow and resource-intensive. Consequently, there is a pressing need for a more general and unified approach that can reliably translate abstract task descriptions into precise control commands, thereby supporting aerial navigation in complex environments.

To address these limitations, data-driven navigation approaches based on Vision Language Action (VLA) \cite{kim2024openvlaopensourcevisionlanguageactionmodel, black2024pi0visionlanguageactionflowmodel} or Vision Language Model (VLM) \cite{bai2025qwen3vltechnicalreport, wang2025internvl35advancingopensourcemultimodal,huanghai}, have emerged as a promising direction for enhancing drone autonomy. By leveraging large-scale pretraining and cross-modal alignment, these models significantly enhance semantic understanding and enable language-conditioned action generation.
With recent advances in model compression techniques such as quantization and knowledge distillation \cite{sink,scaling}, the resulting models have become increasingly acceptable for deployment on onboard platforms. However, most existing VLM/VLA systems are primarily developed for ground-based platforms with fixed viewpoints or for relatively open and simple outdoor environments. As a result, they fall short of meeting the distinctive demands of aerial navigation, particularly in scenarios involving indoor flight, aggressive maneuvers, and tightly constrained spaces. We outline the core challenges inherent in deploying large navigation models on agile drones as follows:

\begin{figure}[tp]
  \centering
  \includegraphics[width=0.98\linewidth]{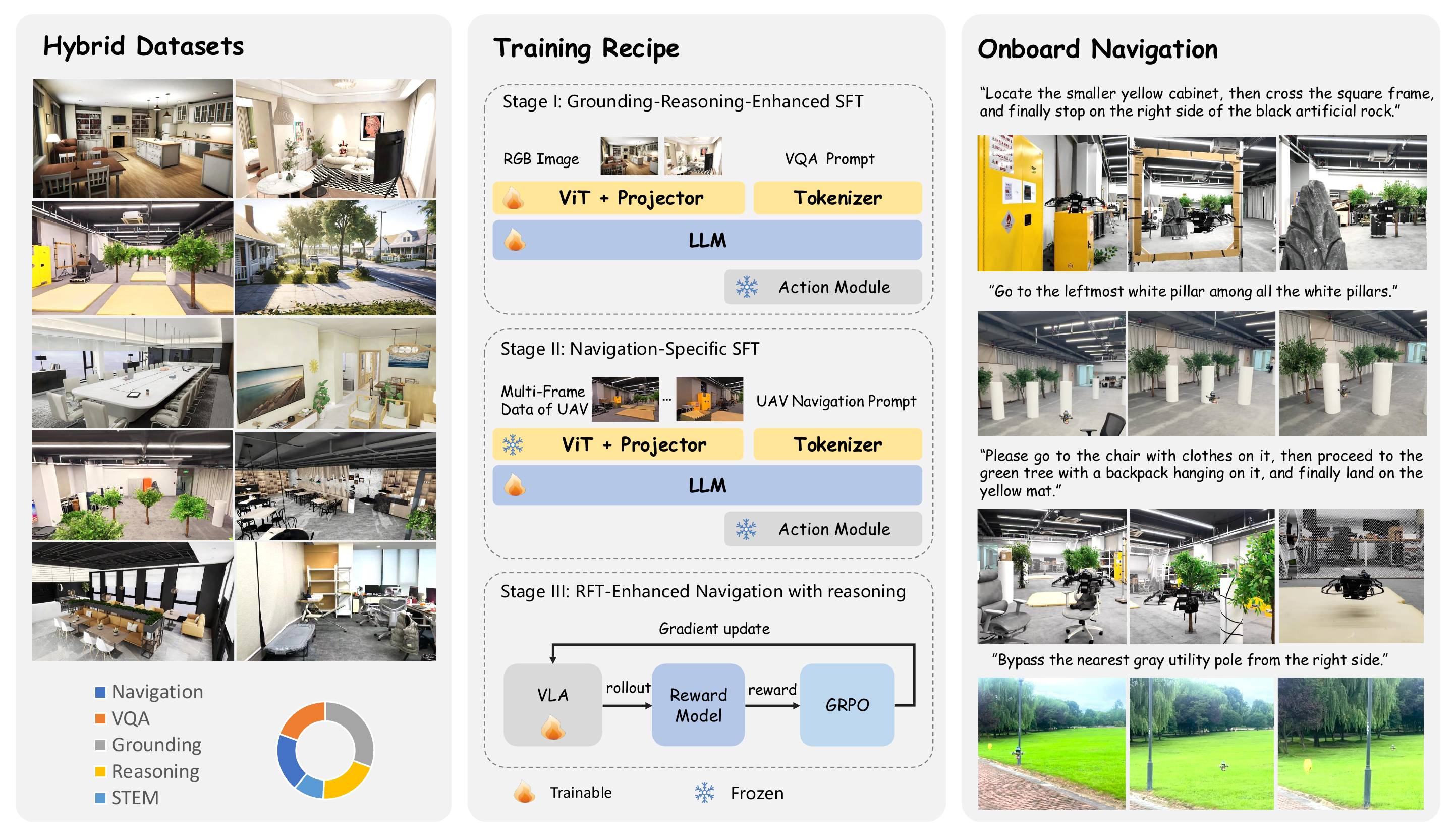}
  \caption{\textbf{Overview of dataset, training, deployment for VLA-AN.} First, VLA-AN constructs a large-scale high-fidelity dataset to bridge the data domain gap. Second, VLA-AN introduces a progressive three-stage training framework to sequentially strengthen spatial grounding, scene reasoning and long-horizon navigation. Third, VLA-AN produces fast, stable, and collision-free action commands that can be executed onboard in real time at 2–3 Hz.}
  \label{fig:fig1}
\end{figure}

\begin{itemize} 
    \item [1)]
    \textbf{Data Distribution Mismatch.} Most large models are pretrained on static images or data captured from fixed or height-constrained viewpoints. This training distribution differs substantially from the highly dynamic, first-person aerial perspective encountered during flight, leading to degraded semantic perception and unstable spatial localization. In addition, acquiring high-quality navigation data from real UAVs is inherently expensive and entails considerable operational risk, further limiting the availability of representative training data.

    \item [2)]
    \textbf{Insufficient Temporal Reasoning for Navigation.} Existing approaches predominantly rely on single-frame inference and have limited ability to encode temporal context. As a result, they struggle to exploit historical observations, reason under complex scenes, or execute long-horizon navigation tasks. These shortcomings hinder stable and continuous operation in real-world environments. Moreover, the reliance on preconstructed or known maps further restricts adaptability when operating in previously unseen spaces.

    \item [3)]
    \textbf{Safety Limitations in Generative Action Models.} Many state-of-the-art VLA models often utilize generative models, such as diffusion policy \cite{chi2023diffusionpolicy} or flow matching \cite{lipman2022flowmatching}, to generate continuous control sequences. While flexible, these approaches introduce stochasticity and generative noise that can significantly increase collision risk, particularly in confined environments. Additionally, these models struggle to incorporate explicit geometric constraints in training collision-free action generation in constrained 3D space.

    \item [4)]
    \textbf{Constraints on Onboard Deployment.} Current VLA models \cite{kim2024openvlaopensourcevisionlanguageactionmodel} impose substantial computational demands, typically requiring high-performance GPUs, which limits their deployment on UAVs with constrained onboard compute and strict payload restrictions. Although approaches such as model distillation \cite{ravichandran2025distillingondevicelanguagemodels} can partially alleviate these constraints, they often incur performance degradation. Consequently, efficient and lightweight frameworks specifically designed for resource-limited UAV deployment remain largely underdeveloped.

\end{itemize}

To systematically address these challenges, we propose an efficient and onboard \textbf{VLA} framework for \textbf{A}erial \textbf{N}avigation (named \textbf{VLA-AN}) in complex environments, establishing a unified pipeline that integrates high-fidelity data generation, multi-stage training, robust action generation, and onboard deployment to achieve full-chain autonomous aerial navigation.

Collecting real-world UAV navigation data is labor-intensive; for instance, UAV-FLOW Colosseo \cite{wang2025uavflowcolosseorealworldbenchmark} spends several weeks to acquire 30K trajectories. Conventional simulators, such as AerialVLN \cite{liu2023aerialvlnvisionandlanguagenavigationuavs}, often lack photorealism, with limited lighting fidelity and texture realism.
To bridge the domain gap between real and synthetic data, we employ 3D Gaussian Splatting (3D-GS) \cite{kerbl2023gaussiansplatting} to construct high-fidelity navigation data. Compared with standard mesh-based rendering, 3D-GS offers superior continuous geometry representation, consistent lighting, and high rendering efficiency, producing synthetic visual scenes that more closely resemble the real world.
Leveraging 3D-GS, we build a large-scale, high-fidelity multimodal navigation dataset that spans diverse indoor and outdoor environments, illumination conditions, occlusion patterns, and dynamic elements. The resulting dataset comprises over 100K navigation trajectories and more than 1M multimodal samples, providing a unified foundation for learning robust semantic navigation across scenes and viewpoints.

Building upon this hybrid dataset, as shown in Fig.~\ref{fig:fig1}, we introduce a progressive three-stage training framework for VLA-AN that systematically enhances navigation with temporal reasoning, particularly long-horizon reasoning, spatial grounding, and fine-grained navigation. Stage I conducts supervised fine-tuning (SFT) on large-scale multimodal data to strengthen scene comprehension, logical reasoning, and spatial inference. Stage II incorporates navigation-centric data to impart essential flight skills, including 3D waypoints generation, yaw prediction, and dynamic re-planning, thereby improving robustness in cluttered environments. Inspired by recent advances in leveraging reinforcement learning to enhance large model capabilities \cite{shao2024deepseekmathpushinglimitsmathematical,li2025simplevlarlscalingvlatraining}, Stage III applies reinforcement learning fine-tuning (RFT) to further refine complex decision-making and enable precise navigation under challenging conditions. Collectively, this framework equips the model to reliably follow complex instructions and perform accurate aerial navigation in real-world scenarios.

A stable action generation module is also essential for reliable navigation. VLA-AN incorporates a real-time robust action module that performs continuous command generation coupled with geometric safety correction. In contrast to large action experts (e.g., the 0.3B flow-matching policy used in $\pi_{0}$ \cite{black2024pi0visionlanguageactionflowmodel}), our design eliminates inference-latency bottlenecks and enables fast, collision-free navigation in dense and previously unseen environments.
Remarkably, VLA-AN can run in real time on lightweight onboard inference platforms, enabled by extensive optimizations throughout our deployment pipeline. These improvements allow the system to sustain inference rates of 2–3 Hz under continuously varying inputs on a 100 TOPS computation board. Our ablation studies further show that the optimized model attains performance comparable to the general pipeline, demonstrating a practical and efficient pathway for real-time onboard deployment on lightweight aerial robots.

\textbf{Contributions.} 
We propose VLA-AN, an efficient and onboard Vision-Language-Action framework for aerial navigation in complex environments. First, we construct a large-scale, high-fidelity 3D-GS dataset for superior realism compared to mesh-based methods, effectively bridging the data domain gap. Second, we introduce a progressive three-stage training framework to sequentially strengthen scene grounding, impart aerial navigation, and enhance complex long-horizon reasoning. Third, to overcome the safety limitations of stochastic generative policies, we design a lightweight, real-time action module coupled with geometric safety correction, ensuring fast, collision-free, and stable command generation without the latency bottlenecks of large action experts. Finally, through extensive system-level optimizations, we enable practical real-time onboard deployment on resource-constrained UAVs, achieving an 8.3× improvement in inference throughput and demonstrating a full-chain closed-loop autonomy pathway for lightweight aerial robots. Extensive simulation and real-world experiments validate its reasoning-enhanced navigation and long-horizon task execution, achieving an average success rate exceeding 90\%.

\section{Related Works}
\label{sec:Realated_Works}

\subsection{Traditional UAV Navigation}

Early UAV autonomous navigation predominantly relied on modular, cascaded architectures following the perception--decision--planning--control paradigm \cite{9102390,10816079,wu2025shapeadaptiveplanningcontroldeformable,9691888}. Localization and environmental perception were typically achieved through LiDAR- or vision-based modules \cite{8421746, 9372856}, while scene semantics were extracted using object detectors such as YOLO \cite{wang2024yolov10}, Faster R-CNN \cite{ren2015faster}, and SceneGraph \cite{chen2025irsinstancelevel3dscene}. High-level task logic was then resolved by rule-based or tree-structured decision modules, followed by motion planners---such as A* \cite{hart1968formal}, RRT \cite{karaman2011sampling}, Fast-Planner \cite{8758904}, and Ego-Planner \cite{9309347}---to generate reference trajectories, which were finally executed by trajectory-tracking controllers to accomplish autonomous flight.

These approaches offer clear structure, strong interpretability, and low deployment cost. However, they are usually tailored for specific task scenarios such as target tracking \cite{9561948}, gap traversal \cite{11128088}, or fast landing \cite{10328688}, but lack the capability to interpret natural-language instructions or leverage the rich semantic information contained in visual observations. Moreover, the serial multi-module pipeline makes the system prone to error accumulation, resulting in notable performance degradation in confined spaces, dynamic environments, and complex three-dimensional spaces.

\subsection{Vision-Language-Action Navigation for Drones}

Recent advances in large-scale language models have enabled a new paradigm for language-driven UAV autonomous navigation, allowing drones to perform cross-modal reasoning between visual semantics and linguistic instructions \cite{hu2025seepointflylearningfree,chen2025gradnavvisionlanguagemodelenabled}. Representative works include AerialVLN \cite{liu2023aerialvlnvisionandlanguagenavigationuavs}, OpenUAV \cite{wang2024realisticuavvisionlanguagenavigation}, CityNav \cite{lee2025citynavlargescaledatasetrealworld}, and more recent frameworks such as FlightGPT \cite{cai2025flightgptgeneralizableinterpretableuav} and NavAgent \cite{liu2024navagentmultiscaleurbanstreet}, which have demonstrated the potential of VLMs in city-scale navigation tasks.

Although existing efforts attempt to construct UAV-specific datasets and achieve instruction alignment (e.g., OpenFly \cite{gao2025openflycomprehensiveplatformaerial}, AVDN \cite{fan2023aerialvisionanddialognavigation}), high-fidelity data with consistent aerial viewpoints remain extremely scarce, constraining the transferability and generalization of  these aerial navigation models. Furthermore, mainstream models are trained primarily on static images or ground-view videos \cite{cheng2025navilaleggedrobotvisionlanguageaction,zhang2025embodiednavigationfoundationmodel}, exhibiting a substantial domain gap from the UAV's high-speed first-person aerial perspective \cite{zhang2025uninavidvideobasedvisionlanguageactionmodel,yu2025thinking360}, which often leads to semantic drift and spatial errors. Most methods rely on single-frame inference and lack temporal modeling capabilities, making them incapable of handling occlusions, historical states, and task progress \cite{hu2025seepointflylearningfree,chen2025gradnavvisionlanguagemodelenabled}---ultimately resulting in behavioral deviations during continuous control. Additionally, many models depend on pre-built maps, limiting their applicability to unknown environments \cite{wang2024realisticuavvisionlanguagenavigation}, and the absence of real-time replanning mechanisms further reduces stability in fast-changing scenes.

\subsection{Onboard Foundation Models for Robotics}

With recent advances in edge computing, deploying foundation models---such as LLMs \cite{openai2023gpt4}, VLMs \cite{bai2025qwen3vltechnicalreport}, and VLA models \cite{kim2024openvlaopensourcevisionlanguageactionmodel}---directly onboard robotic platforms \cite{nvidia2025jetsonthor,nvidia2025jetsonagxorin}, particularly UAVs, has become an emerging trend. Typical approaches utilize quantization, pruning, distillation, and operator fusion to reduce inference cost in order to meet requirements on privacy, safety, and real-time performance \cite{ma2025runningvlasrealtimespeed}. However, it is highly challenging to maintain cross-modal alignment and to achieve low-latency inference on onboard limited computation platforms. For UAV platforms with stringent payload constraints, the model must simultaneously perform visual understanding, language reasoning, and action generation within a unified architecture, imposing extreme demands on computational efficiency and energy consumption.

Existing large-scale robotic models (e.g., the RT models \cite{brohan2023rt2visionlanguageactionmodelstransfer,brohan2023rt1roboticstransformerrealworld}, OpenVLA \cite{kim2024openvlaopensourcevisionlanguageactionmodel}, $\pi_0$ series \cite{black2024pi0visionlanguageactionflowmodel,intelligence2025pi05visionlanguageactionmodelopenworld}) rely heavily on high-performance GPUs and cannot be directly deployed on UAVs. Although recent studies have explored lightweight ViTs \cite{mehta2022mobilevitlightweightgeneralpurposemobilefriendly}, policy distillation \cite{rusu2016policydistillation}, mixed-precision inference \cite{micikevicius2018mixedprecisiontraining}, and hardware --optimized accelerators for onboard deployment, ensuring closed-loop control stability under aggressive quantization remains an unsolved problem. Overall, despite notable progress in onboard foundation models, meeting the stringent accuracy, robustness, and latency requirements under UAV resource constraints remains a major bottleneck. This calls for UAV-oriented efficient model architectures, multimodal compression strategies, and deployable real-time inference frameworks.

\section{Methods}
\label{sec:4}

\begin{figure}
  \centering
  \includegraphics[width=0.98\linewidth]{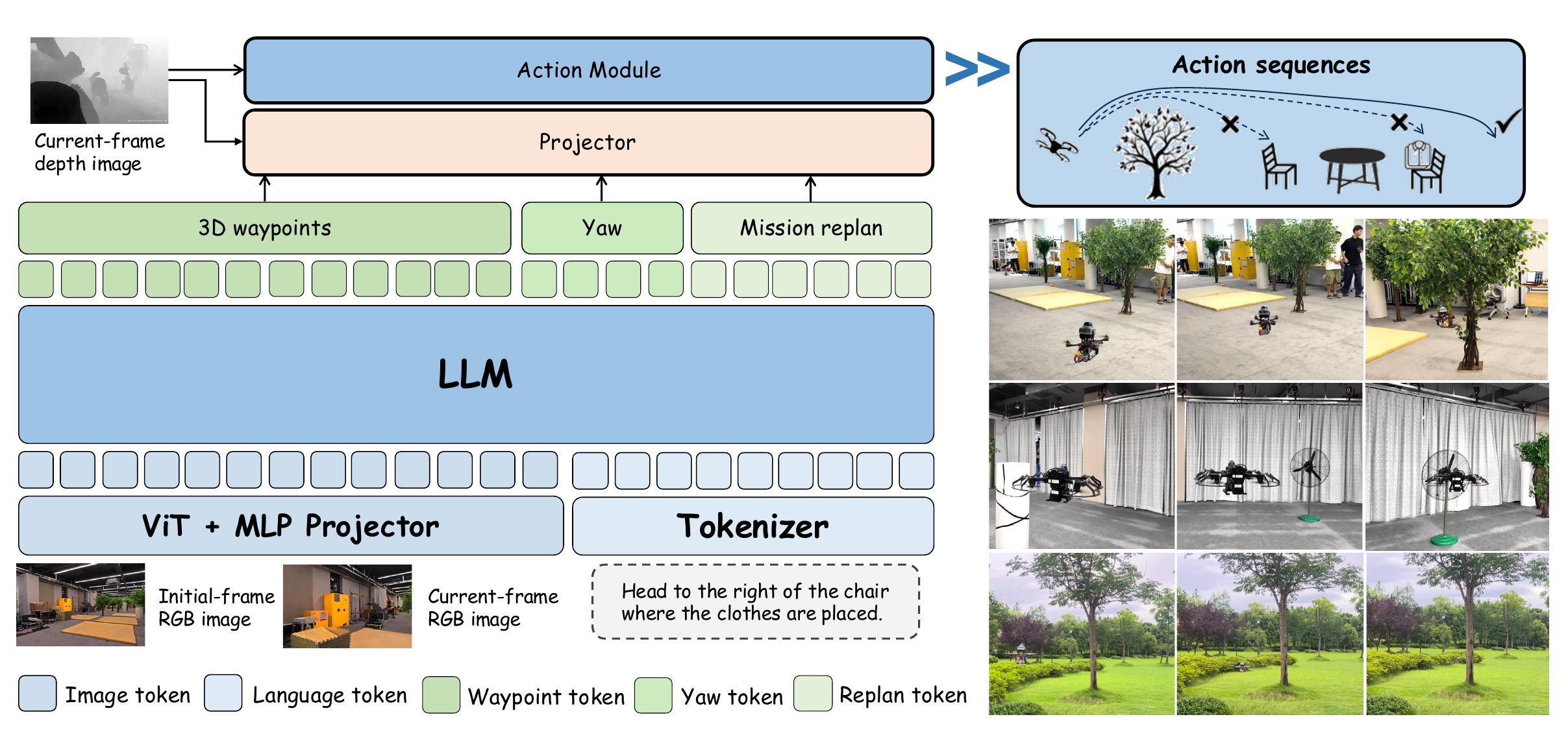}
  \caption{\textbf{Model architecture of VLA-AN.} The VLA-AN model architecture is designed for navigation with reasoning, effectively processing single images and multiple images, and consists of four main components: (1) a Vision Transformer for encoding visual inputs; (2) a MLP projector that maps visual encodings to a latent space aligned with a LLM; (3) the LLM itself for textual understanding and reasoning; and (4) the projector and action module check the desired commands and generate action sequences.}
  \label{fig:fig2}
\end{figure}

As illustrated in Fig.~\ref{fig:fig1}, we construct a hybrid dataset enriched with 3D-GS navigation data to support efficient model training and enhance spatial understanding across diverse viewpoints. The proposed VLA-AN framework adopts a hierarchical architecture that couples semantic reasoning with fine-grained spatial navigation, enabling robust performance on complex, long-horizon tasks under open-vocabulary instructions. Given a natural-language command (e.g., “go to the right side of the white chair with clothes on it”), the model identifies the referenced target entity (the white chair), extracts its attribute constraints (with clothes), parses the corresponding spatial relation (“right side”), and subsequently generates a 3D navigation waypoints along with the desired yaw orientation. By comparing the current visual observation with the initial frame, the model continuously evaluates task progress; if the target has not yet been reached, it initiates a global replanning process to update the navigation goal until successful completion.
To ensure safer and more robust action generation, as shown in Fig.~\ref{fig:fig2}, the projector module and the action module jointly validate and refine the local action sequences, improving obstacle avoidance and guaranteeing secure, efficient UAV maneuvering in cluttered environments. Finally, with extensive optimization for on-device inference, the full VLA-AN system is deployed on a lightweight onboard computing platform, achieving real-time, high-precision autonomous navigation in real-world scenarios.

\subsection{Problem Definition}
\vspace{-0.1cm}

The vision-language navigation task aims to enable an aerial robot to execute autonomous behaviors in unknown environments by interpreting natural language commands and real-time onboard perception. Formally, given a language instruction $L$ and multimodal observations $\mathcal{O}_t = \{I_t^{rgb}, I_t^{depth}, p_t\}$, where $I_t^{rgb}$ and $I_t^{depth}$ denote the RGB and depth images, and $p_t = (x_t, y_t, z_t, \psi_t)$ represents the UAV pose, the objective is to learn a VLA model $\pi_{\text{VLA-AN}}$ that generates an action sequence $a_{1:T}$ maximizing the task success probability $P(\text{success}|L, \mathcal{O}_{1:T})$:

\begin{equation}
\pi_{\text{VLA-AN}} = \arg\max_{\pi} \; \mathbb{E}_{a_{1:T} \sim \pi(\cdot \mid L, \mathcal{O}_{1:T})}
\left[ P\!\left(\text{success} \mid L, \mathcal{O}_{1:T}, a_{1:T}\right) \right]
\end{equation}

Unlike conventional modular pipelines, our framework directly employs the VLA Model to map visual-linguistic inputs to semantic control actions:
\begin{align}
a_t = \pi_{\text{VLA-AN}}(L, I_t^{rgb}, I_t^{depth}).
\end{align}

To maintain task consistency, a temporal comparison module evaluates task completion by contrasting the initial and current frames:
\begin{align}
s_t = f_{\text{cmp}}(I_0^{rgb}, I_t^{rgb}, L),
\end{align}

where $s_t \in \{0,1\}$ denotes the task completion flag. The overall UAV system is formulated as a state transition process:
\begin{align}
& \mathcal{S} = \{ \text{IDLE}, \text{NAVIGATING}, \text{REPLANNING}, \text{TASK\_COMPLETE} \},
\end{align}
ensuring closed-loop autonomous coordination.

This work introduces a comprehensive navigation framework for UAVs in unknown environments, consisting of four core components: a high-fidelity vision-language dataset, a three-stage training procedure, a robust action module, and an onboard deployment framework.

\subsection{High-Fidelity Hybrid Data Collection}
\vspace{-0.1cm}

To bridge the significant domain gap between existing datasets and real-world scenarios regarding visual realism and UAV motion characteristics, we propose an automated data generation pipeline integrating 3D-GS with the Unity engine. This pipeline is designed to generate not only high-fidelity visual observations but also flight trajectories that strictly conform to UAV dynamics. The specific construction process is as follows:

\textbf{Stage I: High-fidelity scene reconstruction via 3D-GS.}
Distinct from traditional mesh-based rendering, we employ 3D-GS technology to capture real-world illumination variations and continuous geometric details. We initially collect video streams covering diverse indoor (e.g., offices, corridors) and outdoor (e.g., parks, streets) environments, converting them into high-quality 3D-GS scene representations before importing them into Unity. The 3D-GS approach preserves real-world visual textures while endowing the simulator with exceptional rendering efficiency and realism, thereby significantly narrowing the sim-to-real visual gap.

\textbf{Stage II: Automated trajectory generation and multimodal data collection.}
To generate large-scale, semantically aligned, and physically constrained navigation data, we establish a closed-loop automated collection system:

\textit{Task Definition and Randomization:} First, human experts formulate the metadata for each subtask, including feasible natural language instructions, trajectory start points, and target endpoints. To enhance data diversity, we perform random sampling of start and end coordinates within reasonable ranges defined by experts, ensuring that tasks remain both challenging and solvable.

\textit{Kinodynamic Planning:} Leveraging the gradient-based trajectory planner, we generate reference trajectories within complex 3D environments. This algorithm fully exploits the spatial maneuverability of UAVs; it not only handles horizontal obstacle avoidance but also flexibly navigates through the Z-axis to evade vertical obstacles, achieving collision-free planning within the full volumetric space.

\textit{Multi-view Synchronous Data Collection:} Within the Unity simulation, the UAV agent flies autonomously along the planned trajectory. The system synchronously records multimodal data at 10 Hz, encompassing RGB images from four views (front, rear, left, right), depth maps, and the UAV's 4D pose in the world coordinate system.

\textbf{Stage III: Hybrid Dataset Construction.}
To strike a balance between sim-to-real visual consistency and environmental diversity in training data, we proportionally blend high-fidelity 3D-GS data with editable mesh data and real-world data. Specifically, 3D-GS scenes effectively bridge the visual domain gap via their photorealism, whereas mesh scenes leverage their high editability to enable flexible adjustments of lighting conditions, occlusion patterns, and target objects. This complementary strategy ensures the model's adaptability to real-world visual features while significantly enhancing robustness against complex and changing environments through domain randomization.










\subsection{Multi-stage Training for VLA}
\vspace{-0.1cm}

To address the coexistence of multiple tasks with significantly different scales, we design a progressive three-stage training framework that systematically enhances cross-task generalization for aerial navigation. This framework integrates multi-frame visual question answering, spatial grounding and reasoning data, as well as high-quality aerial navigation demonstrations. By combining supervised fine-tuning (SFT) with reinforcement learning fine-tuning (RFT), the framework gradually strengthens the model’s spatial understanding, task planning, and high-level reasoning capabilities in complex aerial scenarios, ultimately enabling a unified VLA-based aerial navigation agent.

\textbf{Stage I: Grounding-reasoning-enhanced SFT.}  
In the first stage, we perform full-parameter SFT on a mixture of large-scale general-purpose datasets, including VQA, spatial grounding, reasoning, and STEM tasks, to enhance the model’s fundamental visual understanding, logical reasoning, and spatial relation modeling capabilities. The training corpus further incorporates multi-frame and multi-view image reasoning, temporal consistency modeling across action sequences, and complex aerial-scene analysis. These designs enable the model to expand from object-level perception to scene-level spatial inference, forming a multi-scale understanding mechanism that supports high-level navigation planning. This stage equips the model with essential capabilities such as interpreting natural-language instructions, understanding physical interactions, and parsing real-world spatial layouts, laying a robust foundation for subsequent navigation-specialized training.

\textbf{Stage II: Navigation-specific SFT.}  
The second stage injects high-quality UAV navigation data and proportionally mixes curated VQA reasoning data to construct intensive post-training tailored for aerial navigation. This stage explicitly teaches core UAV skills required in real-world environments, including 3D waypoints planning, desired yaw prediction, and dynamic task re-planning. The model undergoes SFT to maintain broad reasoning abilities inherited from Stage I while adapting to the structured knowledge representation and multi-view observation patterns characteristic of navigation tasks. As a result, the model achieves stronger robustness in autonomous navigation under complex environments and establishes a stable semantic and command-level backbone for the subsequent reinforcement learning stage.

\textbf{Stage III: RFT-enhanced navigation with reasoning.}  
In the third stage, we apply a GRPO-based RFT strategy to further improve accuracy, reliability, and decision consistency. A reward design is constructed from a high-quality thinking dataset to supervise UAV behavior and reasoning processes. For each instruction, multiple responses are sampled, and advantage estimates are computed through within-group normalization. Reward functions are designed to cover visual reasoning, spatial pointing, command generation, etc; visual reasoning is evaluated via strict answer matching, while spatial tasks are assessed using the Intersection over Union (IoU) between predicted and ground-truth bounding boxes or masks. This stage emphasizes correcting failure patterns identified earlier, and under strict format-template validation, the model systematically learns to generate more stable, compliant, and high-confidence outputs. Consequently, the model achieves substantial improvements in executing complex instructions and performing reliable aerial navigation in real-world scenarios.



\subsection{Robust Action Module}
\vspace{-0.1cm}





Existing VLA models rely on generative models to produce action chunking, but inference noise and bias may induce collisions in confined environments, posing significant risks to structurally fragile drones.
To mitigate this, we introduce a robust real-time action module as a continuous command generator to ensure safe and accurate trajectory tracking. 
The module extracts local obstacle information from the depth map only when potential intersections between the trajectory and surrounding obstacles are detected, and generates differentiable repulsive gradient forces that instantly adjust the trajectory. This robust mechanism is computationally efficient and highly responsive, enabling UAVs to achieve low-latency, collision-free navigation in dense, previously unseen environments.

Specifically, the module constructs a reference trajectory from the current state to the target state using onboard sensor data. If the initial trajectory intersects with obstacles, local geometric cues are converted into surface anchors and repulsive directions to refine the flight path and control actions, enabling continuous collision-free adjustment. To preserve dynamic feasibility, the system subsequently re-estimates the temporal allocation and reconstructs the trajectory under safety constraints.
Unlike conventional large generative models—such as the 0.3B flow-matching action expert used in the $\pi_0$ model \cite{black2024pi0visionlanguageactionflowmodel}—the proposed module eliminates the inference-latency bottleneck inherent to such approaches. As a result, the system supports reliable high-speed flight, achieving sustained low-latency, safe, and robust navigation in cluttered and unknown environments.

\subsection{Onboard Deployment Framework}
\vspace{-0.1cm}





Given the strict payload constraints of UAV platforms, most consumer-grade and high-performance computing modules remain impractical for onboard deployment. For instance, NVIDIA Jetson Thor introduces substantial mass relative to typical UAV payload limits, and excessive loading directly reduces flight endurance. Moreover, UAVs are frequently deployed in wide-area, communication-limited, or otherwise challenging environments, where long-duration autonomous operation imposes stringent requirements on onboard inference hardware. Under these constraints, we adopt the lightweight NVIDIA Jetson Orin NX (abbreviated as Orin NX) as our onboard computing unit. After onboard integration and optimization, the total added mass is approximately 80 grams, which is acceptable for micro-scale UAV platforms.

Despite its favorable size–power characteristics, the ARM-based Orin NX offers limited support for the inference of general VLA models, particularly for Vision Transformer (ViT) workloads. In practice, some vision network inference still relies on CPU execution due to a lack of GPU acceleration support, which becomes a major computational bottleneck given the modest CPU performance of Orin NX. Moreover, inference acceleration techniques tailored to VLA models remain nascent; optimizations such as efficient Transformer execution, operator fusion, and KV-cache management have not yet been fully leveraged, preventing effective deployment of VLAs in resource-constrained edge platforms.

To address these challenges, we present a systematic investigation of inference performance optimization for VLA deployment on UAV platforms, with the goal of breaking through the real-time inference bottleneck on limited-resource devices. First, we target multiple acceleration mechanisms across the inference pipeline. Specifically, we integrate a Flash-Attention mechanism, fuse feed-forward network (FFN) layers with normalization operators, and employ a KV-cache pre-loading strategy. These optimizations streamline key computational paths such as memory access, redundant operator execution, and attention computation, thereby substantially reducing latency.

Next, we conduct a fine-grained analysis of computational resource usage across model components, revealing that the ViT module dominates the computational workload. The bottlenecks are twofold: the reliance on CPU for vision network inference, and the ARM architecture’s sub-optimal support for vision operators. To mitigate these, we perform tailored optimizations for ViT’s attention architecture, feature-map dimensionality, and operator scheduling. We restructure operator ordering, optimize memory access patterns, and leverage Single Instruction Multiple Data (SIMD) instruction sets to accelerate execution on ARM.

Furthermore, we incorporate CUDA graph to manage process-level parallelism and pipeline scheduling, reducing inter-process scheduling overhead and enabling the entire VLA inference chain to be executed rapidly within a single process. Collectively, these optimizations significantly enhance the energy efficiency and real-time responsiveness of VLA models on Orin NX, providing a viable deployment path for large models on resource-constrained robotic platforms.

\section{Results}
\label{sec:5}

To comprehensively evaluate the performance of VLA-AN, we conduct extensive experiments and ablation studies focusing on four key questions:
(1) How does VLA-AN compare with state-of-the-art VLA methods?
(2) How well does VLA-AN perform in real-world robotic environments?
(3) What level of performance can be achieved under onboard, resource-constrained deployment?
(4) Are the proposed design components essential and effective?

To achieve this, we conduct comprehensive benchmark tests in simulation, comparing against mainstream VLA methods, followed by extensive real-world experiments to validate the effectiveness and robustness of our approach. Additionally, we evaluate the inference performance of VLA-AN under onboard deployment. Finally, several ablation studies are performed to assess the contribution of key components in our framework.

\begin{figure}
  \centering
  \includegraphics[width=1.0\linewidth]{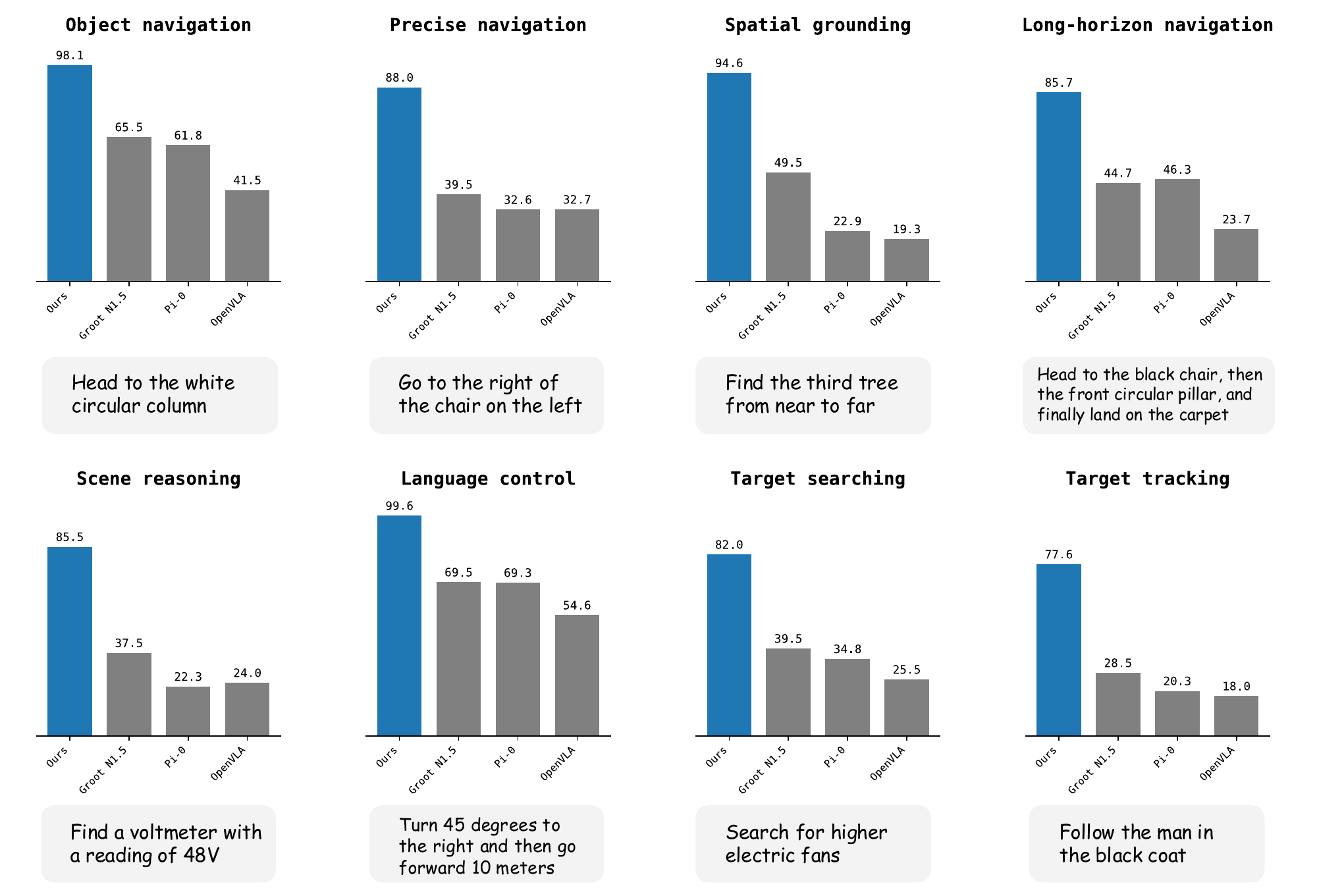}
  \caption{\textbf{Results of benchmark evaluations on diverse navigation tasks using OpenVLA \cite{kim2024openvlaopensourcevisionlanguageactionmodel}, $\pi_{0}$ \cite{black2024pi0visionlanguageactionflowmodel}, Groot N1.5 \cite{nvidia2025groot}, and Ours} (the bar chart values represent the success rate (\%) ).}
  \label{fig:fig3}
\end{figure}

\begin{figure}
  \centering
  \includegraphics[width=1.0\linewidth]{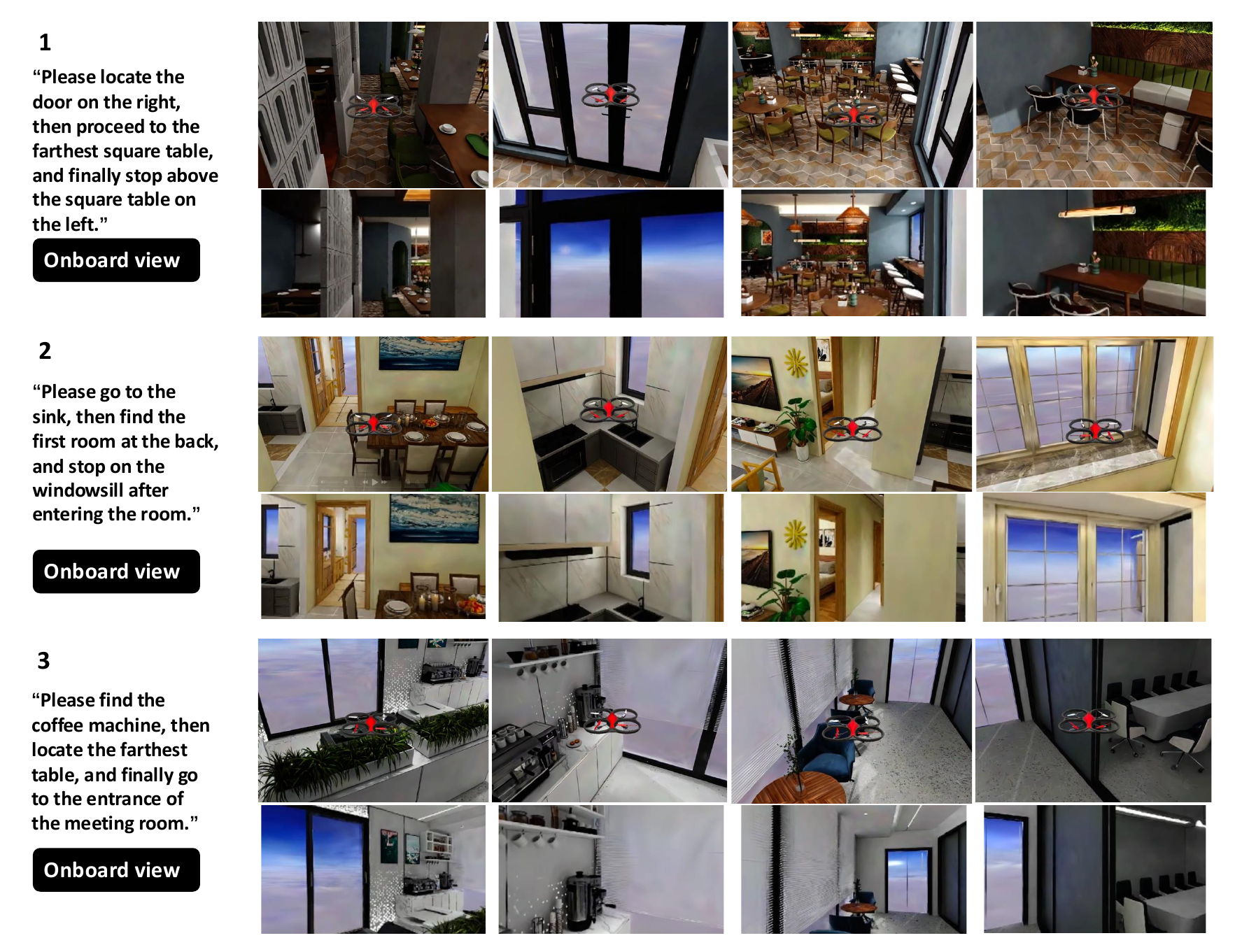}
  \caption{\textbf{Simulation tests for navigation scenarios utilizing both visual and language inputs.}}
  \label{fig:fig4}
\end{figure}

\subsection{Benchmark Results}

We evaluate VLA-AN against several representative VLA baselines, including OpenVLA \cite{kim2024openvlaopensourcevisionlanguageactionmodel}, $\pi_{0}$ \cite{black2024pi0visionlanguageactionflowmodel}, and Groot N1.5 \cite{nvidia2025groot}. Experiments span eight representative navigation scenarios, covering both diverse seen and unseen tasks. Each navigation episode is conditioned on real-time RGB inputs from multiple onboard cameras along with a natural-language instruction. Success Rate (SR) is used as the primary metric. As shown in Fig.~\ref{fig:fig3}, VLA-AN consistently outperforms all baseline models across a wide range of aerial navigation tasks, even achieving an average SR exceeding 98\% for object navigation tasks.

In open-vocabulary object navigation and precise navigation, VLA-AN achieves significant improvements over other models. This advantage arises from the fact that our method does not directly predict low-level motor commands of drones; instead, it outputs target 3D waypoints and desired yaw angles. As a result, the model can learn navigation-specific information from heterogeneous robot datasets (e.g., drones of varying configurations, and even humanoids, quadrupeds), greatly improving the training efficiency of the dataset. This abstraction eliminates the need for precise motor-level predictions and enables a robust action module to apply safe obstacle-avoidance constraints. 
In challenging precise navigation scenarios, such as narrow corridors, cluttered forests, or crowded spaces, models like OpenVLA, $\pi_0$, and Groot N1.5 often produce unsafe motion, leading to numerous failures and a significant decrease in their success rates.

To validate the model's generalization and reasoning capabilities, we also include complex navigation tasks, such as spatial grounding, long-horizon navigation, scene reasoning, language control, and target searching. The test results, shown in Fig.~\ref{fig:fig3}, indicate that VLA-AN significantly outperforms other VLA models across all tasks. This improvement is largely attributed to our multi-stage training strategy, where we inject and enhance substantial reasoning and decision-making capabilities to improve navigation performance in complex scenarios. In contrast, we observe that OpenVLA and $\pi_0$ tend to catastrophically forget their vision-language understanding during task-specific fine-tuning, which degrades multimodal alignment and semantic comprehension. Groot N1.5, due to freezing the VLM weights during training, exhibits stronger reasoning abilities compared to OpenVLA and $\pi_0$.
In spatial reasoning tasks, where the model needs to understand proximity and counting capabilities from multimodal inputs to map to the desired reference state, both OpenVLA and $\pi_0$ achieve relatively low SR. For instance, in the task "Find the third tree from near to far," the model must integrate distance relationships and counting capabilities, which are essential prerequisites for correct action execution.

VLA-AN, surprisingly, also demonstrates strong performance in target tracking, as shown in Fig.~\ref{fig:fig3}. Despite tracking-specific samples constituting less than 1\% of the entire training dataset, the model maintains a stable target tracking capability, achieving an 82.0\% SR on relevant tasks. This robust performance is hypothesized to stem from the fact that our navigation-oriented training implicitly promotes mission re-planning. This mechanism enables the model to autonomously infer whether the target state has changed (e.g., moved or been lost) by aggregating and integrating multi-frame temporal observations. In direct contrast, other existing VLA models exhibit considerably lower SRs in similar dynamic tracking scenarios.

\subsection{Real-World Results}

\begin{figure}[htp]
  \centering
  \includegraphics[width=1.0\linewidth]{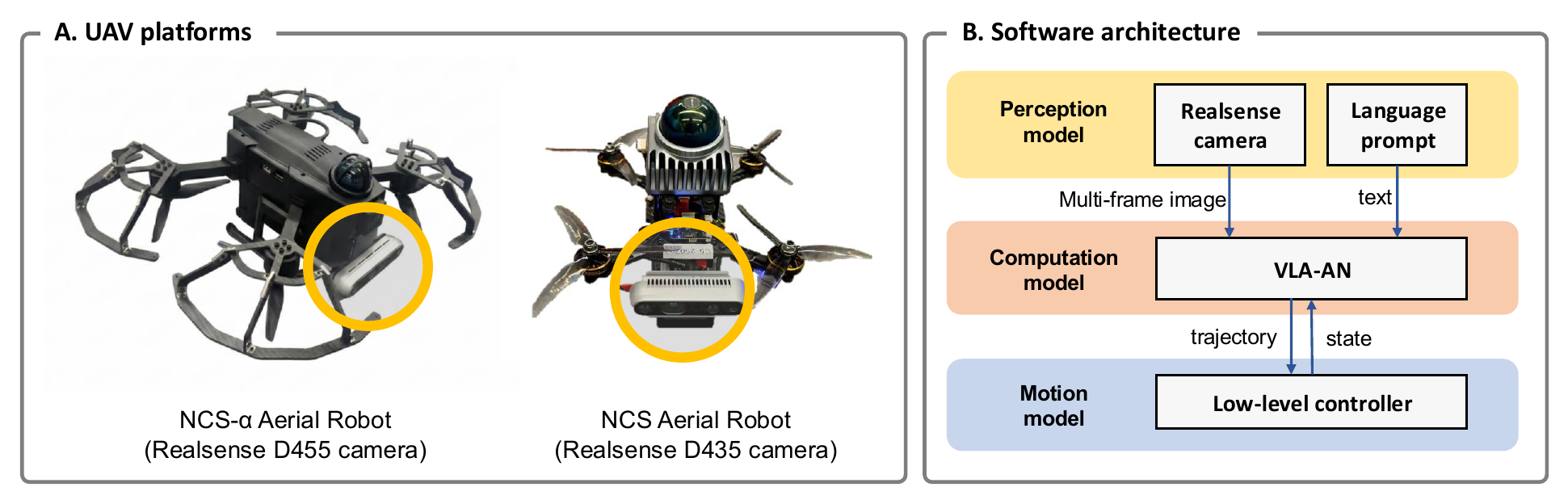}
  \caption{\textbf{Hardware and software architectures of two different experimental drones.}}
  \label{fig:fig5}
\end{figure}

\begin{figure}[htp]
  \centering
  \includegraphics[width=0.97\linewidth]{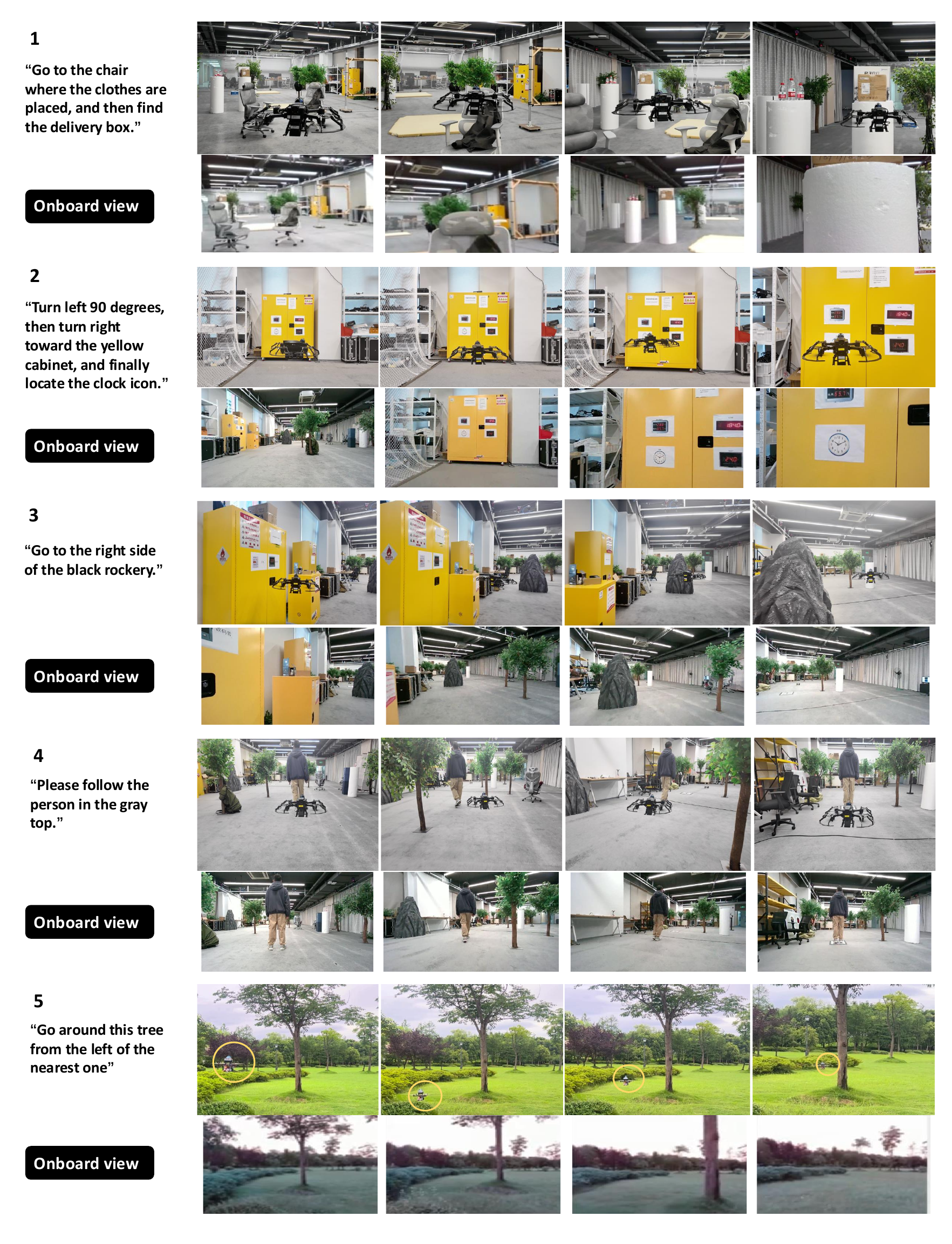}
  \caption{\textbf{Real-world experiments for navigation scenarios utilizing both visual and language inputs.}}
  \label{fig:fig6}
\end{figure}

\textbf{Real-world deployment:}
We deploy VLA-AN on two different UAV platforms, with detailed hardware and software architectures illustrated in Fig.~\ref{fig:fig5}. Each platform is equipped with an Intel RealSense camera for visual perception, and the onboard module performs all perception, language processing, and action generation. During flight, the robots asynchronously transmit the latest observations while executing the previously produced action sequences. In Fig.~\ref{fig:fig6}, we evaluate tasks including object navigation, spatial grounding, scene reasoning, object searching, and target tracking. The success rates remain comparable to those in simulation, demonstrating strong real-world generalization. We further assess long-horizon language instructions, such as “Turn left 90 degrees, then turn right toward the yellow cabinet, and finally locate the clock icon.” VLA-AN correctly decomposes sequential sub-tasks and reliably completes such multi-step instructions.


\textbf{Onboard inference performance:}
As noted earlier, we adopt a lightweight NVIDIA Jetson Orin NX 16 GB module for onboard inference. This lightweight computing board is suitable for small-payload aerial robots and possesses approximately 100 TOPS of computing capability. We deploy VLA-AN models of different scales (7B/3B/2B) under the 30 W performance mode, and the performance results are summarized in Tab.~\ref{tab:onboard_interference}. These results benefit from several system-level optimization strategies, including the integration of flash-attention mechanisms, FFN–Normer operator fusion, KV-cache preloading, and CUDA graph scheduling. Overall, inference throughput improves by 7.1×–9.4× over the unoptimized baselines. During closed-loop navigation, asynchronous image transmission combined with continuous action execution enables high-frequency real-time action inference, which is sufficient for agile autonomous flight in complex environments.

\renewcommand{\arraystretch}{1.5}
\begin{table}[htp]
	\centering	
	\caption{\textbf{Onboard inference performance on different models.}}
	{
		\begin{tabular}{ccc}	
			\hline			
			\textbf{Model} & \textbf{Floating Point Operations} & \textbf{Decode Inference Speed (Second Per Token)} \\ \hline		
			7B-AWQ & $\sim$$~1815.8$ GFLOPs  & 0.110\\  \hline	
            3B-AWQ & $\sim$$~946.7$ GFLOPs  & 0.051 \\ \hline		
			2B-AWQ & $\sim$$~563.3$ GFLOPs  & 0.032 \\ \hline
		\end{tabular}
	}
	\label{tab:onboard_interference}
\end{table}
\renewcommand{\arraystretch}{1}  

Overall, these findings demonstrate that VLA-AN maintains stable and efficient inference performance even under lightweight onboard computing resources, enabling practical real-world deployment of VLA models on drones.

\subsection{Ablation Studies}

\textbf{Different navigation datasets:}
To investigate the effectiveness of 3D-GS data in bridging the domain gap between simulated and real-world visual perception, we evaluate three training configurations: only real-world data, only 3D-GS data, and only mesh data. Notably, the seen dataset contains all types of data, whereas the unseen dataset predominantly consists of real-world data, as our objective is to assess the model’s generalization to real-world navigation. The results are summarized in Tab.~\ref{tab:datasets}. Models trained solely with 3D-GS data achieve performance comparable to those trained on real-world data across both seen and unseen datasets. This can be attributed to 3D-GS providing superior continuous geometry representation, consistent lighting, and efficient rendering, which enhances the realism of synthetic visual features.
In contrast, the model trained exclusively on mesh data exhibits noticeably weaker performance. While it achieves competitive results on the seen dataset, its performance deteriorates on the unseen dataset, which contains more real-world samples, indicating limited generalization to real-world scenarios. Importantly, we adopt a hybrid dataset combining three data types, as it provides a larger and more diverse training set. Although mesh data alone lacks realistic visual fidelity, many open-source datasets are mesh-based, which helps cover additional image viewpoints and further improves navigation success rates.

\renewcommand{\arraystretch}{1.5}
\begin{table}[htp]
    \small
	\centering	
	\caption{\textbf{Ablation study of different training navigation datasets (Success rates \%).}}
	{
		\begin{tabular}{c|ccc|ccc}	
			\hline			
			Datasets 
            & \multicolumn{3}{c|}{Seen Dataset}
            & \multicolumn{3}{c}{Unseen Dataset} \\

            \cline{2-7}
            & \makecell{Object \\ Navigation}
			& \makecell{Long-horizon \\ Navigation}
            & \makecell{Spatial \\ Grounding}
            & \makecell{Object \\ Navigation}
			& \makecell{Long-horizon \\ Navigation}
            & \makecell{Spatial \\ Grounding} \\ 
            \hline	
		
			Only real-world data 
			& 98.8 & \textbf{89.1} & 95.0
            & 96.3 & 82.7 & 90.4 \\ \hline

			Only 3D-GS data 
			& 98.6 & 87.4 & 95.2
            & 95.8 & \textbf{82.9} & 91.1 \\ \hline

			Only mesh data 
			& 92.2 & 78.0 & 88.3
            & 70.3 & 61.8 & 74.2 \\ \hline

            Hybrid data 
			& \textbf{99.2} & 88.2 & \textbf{96.5}
            & \textbf{97.6} & 83.5 & \textbf{92.8} \\ \hline
		\end{tabular}
	}
	\label{tab:datasets}	
\end{table}
\renewcommand{\arraystretch}{1}

\textbf{Multi-stage training strategy:}
To validate the effectiveness of the proposed progressive three-stage training framework for multi-task aerial navigation, we conduct systematic ablation studies (Tab.~\ref{tab:training}). The results indicate that different training stages contribute distinctly to various task performances. Stage I (VLA-AN w/ reasoning) significantly enhances scene reasoning and spatial grounding, achieving 87.2\% and 94.5\% success rates, respectively, demonstrating that early full-parameter SFT effectively strengthens the model’s visual grounding and logical reasoning capabilities. Stage II alone (VLA-AN w/ navigation) excels in target and long-horizon navigation tasks, with success rates of 95.3\% and 82.9\%, respectively, indicating that navigation-specific subsequent training markedly improves path planning and action execution. When Stage I and Stage II are combined (VLA-AN w/ full SFT), high performance is maintained across all tasks, confirming that multi-stage training simultaneously enhances visual reasoning and navigation proficiency. The introduction of Stage III reinforcement learning fine-tuning (VLA-AN w/ RFT) further improves long-range navigation success, highlighting RFT’s potential in correcting navigation failure patterns and optimizing decision consistency. Ultimately, the fully integrated three-stage model (Ours) achieves the superior performance across the majority of tasks: scene reasoning 85.5\%, spatial localization 94.6\%, target navigation 98.1\%, and long-range navigation 85.7\%, fully demonstrating the effectiveness and necessity of the progressive multi-stage framework for cross-task generalization.
These results suggest that by combining our progressive three-stage training, the model can perform robustly and efficiently in complex aerial environments.

\renewcommand{\arraystretch}{1.5}
\begin{table}[htp]
    \centering
    \small
    \setlength{\tabcolsep}{3pt}
    \caption{\textbf{Ablation study of multi-stage training strategy (Success rates \%).}}
    {
        \begin{tabular}{cccccccc}
            \hline
            Model & \makecell{Stage I} & \makecell{Stage II} & 
            \makecell{Stage III}
            & \makecell{Scene\\Reasoning}
            & \makecell{Spatial\\Grounding}
            & \makecell{Object\\Navigation}
            & \makecell{Long-horizon\\Navigation} \\ \hline
            Baseline & $\times$ & $\times$ & $\times$ & 77.4 & 68.2 & 34.2 & 20.4 \\ \hline
            VLA-AN w/ reasoning & $\checkmark$ & $\times$ & $\times$ & \textbf{87.2} & 94.5 & 38.1 & 22.5 \\ \hline
            VLA-AN w/ navigation & $\times$ & $\checkmark$ & $\times$ & 75.6 & 64.3 & 95.3 & 74.9 \\ \hline
            VLA-AN w/ full SFT & $\checkmark$ & $\checkmark$ & $\times$  & 82.3 & 91.1 & 92.9 & 72.2 \\ \hline
            VLA-AN w/ RFT & $\times$ & $\times$ & $\checkmark$ & 82.6 & 76.1 & 45.4 & 36.3 \\ \hline
            Ours & $\checkmark$ & $\checkmark$ & $\checkmark$ & 85.5 & \textbf{94.6} & \textbf{98.1} & \textbf{85.7} \\ \hline
        \end{tabular}
    }
    \label{tab:training}
\end{table}
\renewcommand{\arraystretch}{1}

\begin{figure}[htp]
  \centering
  \includegraphics[width=0.9\linewidth]{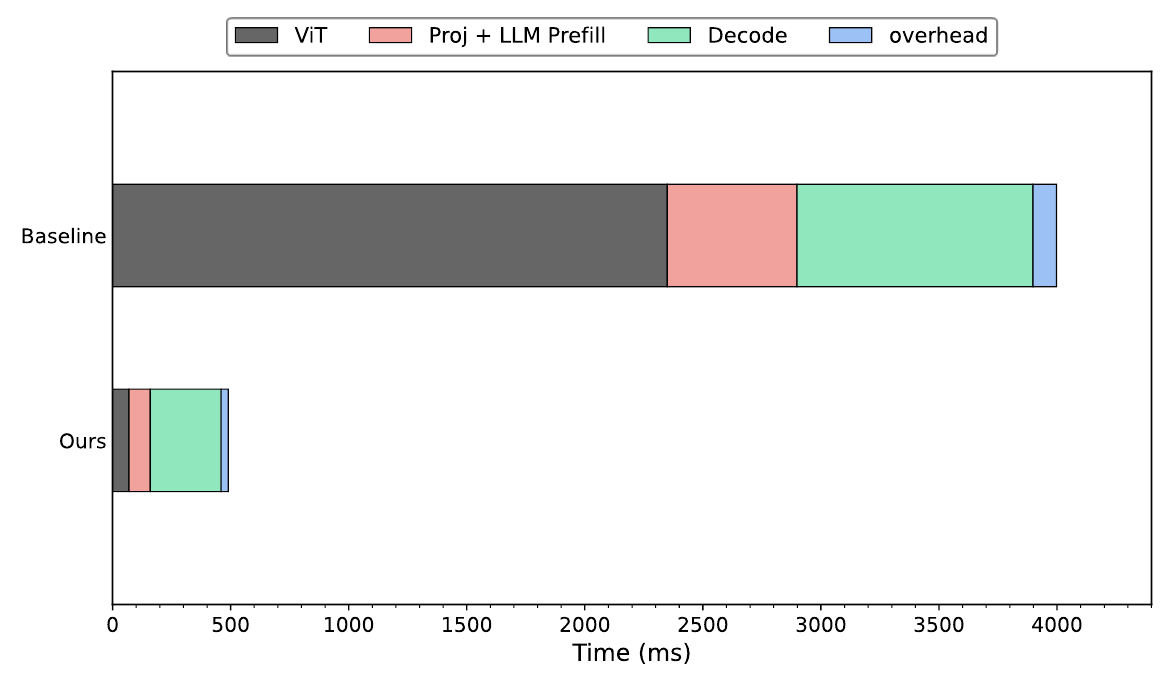}
  \caption{\textbf{Ablation study of onboard inference optimization.} The figure illustrates the original inference time and the inference time achieved after applying various optimization techniques.}
  \label{fig:fig7}
\end{figure}

\textbf{Onboard inference optimization:}
Finally, we benchmark our optimized method on the onboard NVIDIA Jetson Orin NX under ARM architecture against several popular frameworks. Specifically, we employ multiple system-level optimization strategies, including flash-attention mechanisms, FFN–(RMS)Norm fusion, KV-cache preloading, and CUDA graph scheduling. For detailed inference evaluation, we use 720P relatively high-resolution images as visual input and variable-length, non-empty textual prompts (near 50 tokens on average), reflecting interactive onboard usage. The performance is compared against the unoptimized baseline, and Fig.~\ref{fig:fig7} illustrates the inference time improvements contributed by each optimization. Notably, the total inference time is reduced from about 4100 ms to 494 ms (8.3× speedup, 87.9\% reduction). The largest absolute reduction occurs in the ViT stage (2350 → 120 ms), while both projector + LLM prefill (550 → 130 ms) and decoding (1000 → 230 ms) are substantially accelerated, together enabling real-time capable onboard deployment under constrained computational resources. In addition, we apply pipelined execution to the CPU-intensive action module, effectively overlapping computation and control logic to further reduce end-to-end latency on the ARM-based onboard platform.

\section{Conclusion}
\label{sec:6}

This work presents VLA-AN, an integrated framework that advances aerial robot autonomy by combining high-fidelity data generation, multimodal learning, safety-aware action generation, and efficient onboard deployment. Using 3D-GS method, we build a realistic UAV-centric dataset that narrows the domain gap between synthetic and real-world observations. The three-stage training paradigm endows the model with strong scene comprehension, temporal reasoning, and long-horizon navigation, while the lightweight real-time action module ensures safe and collision-free control in constrained, unseen environments. Experiments show that VLA-AN significantly improves semantic grounding, trajectory robustness, and cross-scene generalization, achieving reliable performance even on resource-limited platforms, offering a pathway toward scalable and cognitively capable aerial agents for diverse UAV applications.

Looking forward, we aim to expand the aerial navigation dataset to capture richer UAV observations and actions across varying altitudes, thereby enhancing 3D spatial navigation and reasoning capabilities. We also plan to further optimize inference speed using more advanced techniques and higher-performance hardware (e.g., NVIDIA Jetson Thor) to approach real-time inference rates near 10 Hz, meeting the demands of responsive decision-making. Additionally, we intend to develop navigation-adapted attention mechanisms to train a more intelligent flight foundation navigation model capable of sophisticated perception and action in complex environments.

\newpage

\bibliography{references}

\bibliographystyle{sciencemag}

\end{document}